\documentclass{./styles/svproc}
\usepackage{url}
\usepackage{graphicx}
\usepackage{float}

\usepackage[margin=1in]{geometry}
\usepackage{amsmath}
\usepackage{amssymb}
\usepackage{booktabs}
\usepackage{multirow}
\usepackage{algorithm}
\usepackage{algorithmic}
\usepackage{authblk}
\usepackage{subcaption}
\usepackage{cleveref}

\newcommand{\framework}{LLMForge}
\newcommand{\pipelineA}{IterTracer}
\newcommand{\pipelineB}{IterVision}

\begin{document}
\mainmatter              

\title{Foundation Models for Automatic CAD Generation}

\titlerunning{Foundation Models for Automatic CAD Generation}

\author{J. de Curt\`o\inst{1,2}, Victoria Guill\'en\inst{2,3} \and I. de Zarz\`a\inst{4}}

\authorrunning{de Curt\`o et al.}

\tocauthor{de Curt\`o et al.}

\institute{
Department of Computer Applications in Science \& Engineering, BARCELONA Supercomputing Center, \\08034 Barcelona, Spain 
\and 
Escuela Técnica Superior de Ingeniería (ICAI), Universidad Pontificia Comillas, 28015 Madrid, Spain
\and
Department of Artificial Intelligence, Chung-Ang University, Seoul, Republic of Korea
\and Human Centered AI, Data \& Software, LUXEMBOURG Institute of Science and Technology, L-4362 Esch-sur-Alzette, Luxembourg}

\maketitle              

\begin{abstract}

Recent advances in Large Language Models~(LLMs) and
Vision-Language Models~(VLMs) have opened new pathways for the
automatic generation of parametric three-dimensional designs from
natural-language specifications.
This chapter presents a comprehensive empirical study on the use of
modern foundation models for automatic Computer-Aided Design~(CAD)
generation of mechanical parts, structured around a unified evaluation
pipeline and a curated benchmark suite of~97 engineering design problems.
We introduce \framework{}, a multi-model text-to-CAD framework that
integrates JSON-schema validation, analytic feature scoring, mesh
synthesis, and a multi-round iterative refinement loop, studied under
two distinct critique regimes.
The first regime, \pipelineA{}, employs a Phong-shaded ray-trace
renderer coupled with a suite of analytic visual metrics,
silhouette~IoU, hole visibility, edge clearance, and aspect-ratio
conformance, to provide lightweight, geometry-aware feedback across
successive generation rounds.
The second regime, \pipelineB{}, replaces the analytic scorer with a
VLM-based semantic critic~(Qwen2.5-VL-72B) that evaluates rendered
views of each candidate geometry through chain-of-thought visual
reasoning, enabling richer assessment of spatial coherence and design
intent.
Using a benchmark spanning four canonical geometry families,
rectangular plates with holes and bolt circles, multi-feature boxes,
flanged cylinders, and L-brackets, we evaluate seven state-of-the-art
foundation models: DeepSeek-V3.2, Qwen3-235B-A22B, Llama-3.3-70B,
Gemma-3-27B, GLM-4.5, MiniMax-M2.1, and INTELLECT.
Under \pipelineA{}, the four highest-ranked models form a tight
performance cluster~($\mu_{\text{overall}}\in[0.885,\,0.890]$) with mesh
success rates of~98.97\%, demonstrating that compact instruction-tuned
models can attain reliability competitive with substantially larger
systems.
The addition of VLM-based critique in \pipelineB{} yields 100\%
watertight mesh generation on the leading model while surfacing
systematic difficulty on rotationally symmetric geometries such as
cylinders, where visual and semantic scoring diverge most markedly.
The chapter discusses benchmark design principles, model failure modes,
CAD-oriented prompting strategies, and implications for industrial
engineering workflows.
We conclude by identifying future research directions for scalable,
automated mechanical design within Global Applied AI pipelines.

\keywords{Large Language Models,
          Vision-Language Models,
          Text-to-CAD,
          Parametric Design,
          Foundation Models,
          Engineering Automation}

\end{abstract}

\section{Introduction}
\label{sn:introduction}
 
The automation of mechanical design has long been a central aspiration
of engineering informatics.
Traditional Computer-Aided Design~(CAD) workflows demand expert
knowledge of constraint-based parametric modelling, scripting
languages, and domain conventions that are largely inaccessible to
non-specialists and difficult to integrate into automated processing
chains.
The emergence of large language models capable of generating structured
code from natural-language instructions represents a qualitative shift
in this landscape: a practitioner can now describe a mechanical part in
plain English,  \textit{``a 150\,mm $\times$ 100\,mm $\times$ 4\,mm
plate with a centred bolt circle of six equally-spaced M5 holes,
diameter~60\,mm''},  and expect a computational agent to translate that
intent directly into a valid, manufacturable three-dimensional geometry.
 
This chapter reports on \framework{} (\textbf{L}anguage-\textbf{L}arge-\textbf{M}odel
\textbf{F}oundation-driven \textbf{O}ptimised \textbf{R}efinement
\textbf{G}eneration \textbf{E}ngine), a systematic study of this
capability across seven state-of-the-art foundation models examined
under two progressively richer evaluation regimes.
The work addresses a practical need that is increasingly common across
precision manufacturing and aerospace engineering contexts: large numbers
of structural components must be regenerated or adapted from engineering
specifications with limited manual intervention.
Automating this process through language-guided generation and
multi-round iterative critique could substantially reduce design
iteration cycles and lower the barrier to rapid prototyping.
 
The central challenge is not merely generating syntactically valid code,
but generating code that is \emph{semantically correct} with respect to
geometric intent.
A CAD script that parses without error may nonetheless produce a
geometry with incorrect dimensions, missing features, or topological
defects that prevent downstream meshing or simulation.
This motivates the design of \framework{}'s multi-dimensional evaluation
pipeline, which scores generated parts along four independent axes, 
schema validation, mesh soundness, feature adherence, and visual
fidelity,  and iterates the generation loop up to four rounds per
problem, allowing models to self-correct in response to structured
feedback.
 
\subsection*{Contributions}
 
This chapter makes the following specific contributions.
 
\begin{enumerate}
 
  \item \textbf{\pipelineA{}}: an iterative text-to-CAD generation
    pipeline in which each candidate geometry is rendered via a
    Phong-shaded ray tracer and scored against a suite of analytic
    visual metrics~(silhouette~IoU, hole visibility, edge clearance,
    aspect-ratio conformance). The resulting scalar critique is
    incorporated into the next-round prompt, enabling lightweight
    geometry-aware feedback without a secondary neural model.
    (\Cref{sn:pipelines})
 
  \item \textbf{\pipelineB{}}: an extension of the pipeline that
    replaces the analytic visual scorer with a VLM-based semantic
    critic~(Qwen2.5-VL-72B operating in chain-of-thought mode). This
    enables higher-level assessment of spatial coherence and design
    intent, and introduces a new \emph{VLM score} dimension into the
    evaluation record alongside validation, mesh, feature, and visual
    axes. (\Cref{sn:pipelines})
 
  \item \textbf{A benchmark of 97 engineering design problems} spanning
    four canonical geometry families: rectangular plates with
    holes and bolt circles, multi-feature boxes, flanged cylinders, and
    L-brackets. Each problem is paired with a ground-truth feature
    specification used to compute analytic feature adherence scores.
    (\Cref{sn:benchmark})
 
  \item \textbf{A comparative empirical evaluation of seven foundation
    models},  DeepSeek-V3.2, Qwen3-235B-A22B, Llama-3.3-70B,
    Gemma-3-27B, GLM-4.5, MiniMax-M2.1, and INTELLECT,  across both
    pipelines, producing 679 scored design attempts per system with
    per-round convergence trajectories and full mesh-quality statistics.
    (\Cref{sn:results})
 
\end{enumerate}
 
\subsection*{Key Findings}
 
Under the \pipelineA{} regime, four models~(DeepSeek-V3.2,
Qwen3-235B-A22B, Llama-3.3-70B, and Gemma-3-27B) achieve overall scores
within a remarkably tight band of $[0.885,\,0.890]$, each with a mesh
success rate of 98.97\%, suggesting that the text-to-CAD task for this
class of problems has reached a saturation point for top-tier instruction-tuned
models under analytic visual critique.
The three remaining models~(GLM-4.5, MiniMax-M2.1, INTELLECT) exhibit
substantially higher variance~($\sigma > 0.28$) and lower success
rates, reflecting qualitatively different failure modes: schema
non-conformance, degenerate meshes, and persistent feature omission.
 
Under the \pipelineB{} regime, the addition of VLM-based critique
exerts a more demanding semantic assessment pressure, reflected in a
uniform reduction of approximately 0.04 points in overall score across
the top-four cluster compared to~\pipelineA{}.
Despite this, Gemma-3-27B achieves a mesh success rate of
\emph{100\%}~(97/97), and all 512 analysed meshes are fully watertight
and topologically valid solid volumes.
Cylinder geometries emerge as the most challenging category across all
models, with visual and VLM scores diverging by up to 0.15 points
relative to plates and L-brackets, indicating a systematic difficulty
with rotationally symmetric features that analytic metrics partially
overlook.
 
\subsection*{Scope and Limitations}
 
The benchmark is intentionally bounded to four canonical geometry
families to enable systematic comparison; more complex assemblies,
freeform surfaces, and multi-body parts lie outside the current scope.
All experiments are conducted using Nebius AI Studio as the inference
backend, with temperature settings held constant across models to
ensure comparability.
The VLM scorer in \pipelineB{} introduces stochastic evaluation
variance that is partially absorbed by the multi-round protocol but
cannot be fully eliminated; we report VLM scores descriptively and
treat them as complementary to, rather than replacements for, the
analytic metrics.
 
\subsection*{Chapter Organisation}
 
The remainder of this chapter is organised as follows.
\Cref{sn:related} reviews related work on LLM-assisted design, code
generation for CAD, and VLM-based quality assessment.
\Cref{sn:benchmark} presents the benchmark dataset and evaluation
pipeline in detail, covering problem construction, the JSON-schema CAD
representation, scoring components, and the iterative refinement
protocol.
\Cref{sn:pipelines} describes the two critique regimes, 
\pipelineA{} and \pipelineB{},  and their respective signal
characteristics.
\Cref{sn:results} reports experimental results, including per-model
rankings, part-type difficulty analysis, round-level convergence,
mesh-quality statistics, and a comparative analysis between the two
systems.
\Cref{sn:discussion} discusses model failure modes, the effect of
model scale and instruction tuning, and the implications for deploying
foundation models in industrial CAD pipelines.
\Cref{sn:conclusion} concludes with a research outlook toward
scalable, automated mechanical design.

\section{Related Work}
\label{sn:related}
 
\paragraph{Foundation models and code generation.}
The transformer~\cite{vaswani2017} and the demonstration of few-shot
generalisation in large autoregressive models~\cite{brown2020} underpin
all modern code-generating systems.
Codex~\cite{chen2021codex} established that models trained on code
corpora can complete function bodies from natural-language docstrings,
motivating subsequent work on program synthesis~\cite{austin2021} and
multi-turn generation~\cite{nijkamp2023codegen}.
General-purpose instruction-tuned models,  including
GPT-4~\cite{openai2023gpt4} and the LLaMA family~\cite{touvron2023llama,grattafiori2024llama3}, have since narrowed the gap with specialist code models,
making them practical backends for structured-output tasks
such as the JSON-schema CAD scripts used in \framework{}.
Iterative self-correction via structured feedback has been shown to
improve generation quality without retraining~\cite{madaan2023selfrefine},
and LLMs can leverage execution error signals to self-debug across
multiple correction rounds~\cite{chen2024intervenor}.
These results directly motivate the multi-round refinement loop in
both \pipelineA{} and \pipelineB{}.
 
\paragraph{Generative models for CAD.}
Parametric CAD generation,  producing sequences of modelling
operations rather than raw geometry,  has attracted growing
attention.
DeepCAD~\cite{wu2021deepcad} proposed a transformer-based autoregressive
model for CAD construction sequences, and SkexGen~\cite{xu2022skexgen}
improved generalisability by disentangling topology and geometry into
separate codebooks.
\cite{makatura2023llm} demonstrated that
general-purpose LLMs such as GPT-4 can assist in parametric design
tasks through interactive dialogue, positioning the LLM-prompting
approach as a practical complement to dedicated architectures.
\framework{} extends this line by systematically comparing seven
off-the-shelf instruction-tuned models under a unified, reproducible
evaluation pipeline across four canonical geometry families.
 
\paragraph{Vision-language models and model-as-judge evaluation.}
Flamingo~\cite{alayrac2022}, LLaVA~\cite{liu2023llava},
CogVLM~\cite{wang2025}, and Qwen2-VL~\cite{qwen2vl} have
demonstrated that fusing visual and linguistic representations enables
rich cross-modal reasoning.
GPT-4V~\cite{openai2023gpt4} further showed that such models can act as
surrogate evaluators of visual artefacts.
Surveys of LLM evaluation~\cite{chang2024survey} and
code-generation benchmarking~\cite{zhuo2024} have highlighted the
importance of decomposed, multi-axis scoring over binary pass/fail
verdicts,  a principle that \framework{} applies to the CAD domain.
The \pipelineB{} pipeline instantiates the model-as-judge paradigm for
parametric CAD by deploying a Qwen2.5-VL-72B critic within the
iterative refinement loop, building on earlier LLM applications to
engineering reasoning~\cite{deCurto2025_2,deCurto2025_3}.

\section{Benchmark and Evaluation Pipeline}
\label{sn:benchmark}
 
 
The benchmark consists of 97 engineering design problems encoded as
natural-language strings, drawn from four canonical geometry families:
rectangular \emph{plates} with holes and bolt circles, hollow and solid
\emph{boxes}, \emph{cylinders}, and \emph{L-brackets}.
Problems span a range of specification complexity, from a plain
\textit{``Make a simple 100$\times$80$\times$5\,mm rectangular plate''} to
multi-feature descriptions combining bolt circles, individual hole
offsets, chamfers, and fillets.
Each problem record carries a \texttt{part\_type} label and a
\texttt{difficulty} tag used for stratified analysis; ground-truth
feature specifications (hole count, nominal dimensions, and surface
treatments) are stored alongside each problem and used to compute the
analytic feature adherence score described in
\Cref{sn:benchmark:scoring}.
 
 
Foundation models are prompted to emit a strictly structured JSON
object rather than free-form geometry code.
The schema, held constant across all models and both pipelines, is:
 
\begin{verbatim}
{
  "part_type"    : "plate"|"box"|"cylinder"|"l_bracket",
  "units"        : "mm",
  "params"       : { ... },
  "holes"        : [{"x":n,"y":n,"diameter":n,"pattern_id":s}],
  "fillet_radius": number,
  "chamfer"      : number,
  "transform"    : {"rx":0,"ry":0,"rz":0,"tx":0,"ty":0,"tz":0}
}
\end{verbatim}
 
The \texttt{params} block is keyed on \texttt{part\_type}:
\texttt{plate} requires \texttt{width}, \texttt{height}, and
\texttt{thickness}; \texttt{box} adds \texttt{depth} and an optional
\texttt{thickness} (null for a solid box); \texttt{cylinder} requires
\texttt{diameter} and \texttt{height}; \texttt{l\_bracket} requires leg
dimensions \texttt{a}, \texttt{b}, \texttt{thickness}, and
\texttt{leg\_width}.
All quantities are in millimetres.
The initial system prompt enforces metric-to-clearance mappings for
threaded holes (e.g.\ M3$\to$3.4\,mm, M4$\to$4.5\,mm, M5$\to$5.5\,mm,
M6$\to$6.6\,mm, M8$\to$9.0\,mm) and requires hole coordinates to be
expressed relative to the part centre.
Models must return the JSON object only, with no markdown fences or
surrounding text; a regex-guarded parser strips residual formatting
before further processing.
 
 
A deterministic Python geometry engine converts each valid JSON spec
into a watertight triangulated mesh using
Trimesh and Shapely.
For plates, a 2-D rectangular Shapely polygon is constructed
at the nominal dimensions and each hole is subtracted as a
96-segment circular buffer at the declared centre and clearance
diameter; the resulting polygon (resolved through a
\texttt{buffer(0)} validity repair if needed) is then extruded to the
declared thickness.
For boxes, a solid box primitive is created with Trimesh's
\texttt{creation.box}; if a wall thickness is specified, an inner box
is subtracted via Boolean difference.
Cylinders are tessellated with 96 circumferential sections.
L-brackets are assembled from two box primitives: a
horizontal base leg and a vertical leg obtained by rotating the second
box $90^{\circ}$ about the $y$-axis, then concatenating both into a
single mesh object.
After construction, degenerate faces, unreferenced vertices, and
duplicate vertices are removed.
 
\subsection{Scoring Axes}
\label{sn:benchmark:scoring}
 
Each generated part is evaluated along four independent axes.
 
\textbf{Validation score} ($s_\text{val} \in [0,1]$) measures JSON
schema conformance: presence of all required keys, correct
\texttt{part\_type} value, numeric types for all parameters, and
absence of structurally invalid entries.
 
\textbf{Mesh score} ($s_\text{mesh} \in \{0,1\}$) is a binary
indicator that equals 1 if and only if the geometry engine produces a
non-empty mesh without raising an exception, and 0 otherwise.
 
\textbf{Feature score} ($s_\text{feat} \in [0,1]$) compares
the generated specification against the ground-truth feature record:
hole count, principal dimensions, and declared surface treatments
(chamfer, fillet) are checked analytically, and partial credit is
awarded proportionally.
 
\textbf{Visual score} ($s_\text{vis} \in [0,1]$) is a composite of
five sub-signals derived from rendered bitmaps of the candidate mesh:
silhouette IoU (fill ratio of part pixels within the tight bounding
box of the top-view projection), hole visibility (fraction of declared
holes whose projections are visible from the top-view render), edge
clearance (minimum distance from any hole projection to the plate
boundary), aspect-ratio conformance, and a cross-sectional consistency
score.
The sub-signal weights are: silhouette IoU 0.30, hole visibility 0.30,
edge clearance 0.20, aspect ratio 0.10, section score 0.10.
 
The composite overall score differs between the two pipelines.
Under \pipelineA{}:
\begin{equation}
  s_\text{overall}^{\text{A}} =
    0.25\,s_\text{val} + 0.15\,s_\text{mesh} +
    0.20\,s_\text{feat} + 0.40\,s_\text{vis}.
  \label{eq:score_v35}
\end{equation}
Under \pipelineB{}, a fifth axis,  the VLM semantic match
$s_\text{vlm}$ returned by the Qwen2.5-VL-72B critic,  is
incorporated:
\begin{equation}
  s_\text{overall}^{\text{B}} =
    0.20\,s_\text{val} + 0.10\,s_\text{mesh} +
    0.20\,s_\text{feat} + 0.30\,s_\text{vis} + 0.20\,s_\text{vlm}.
  \label{eq:score_v40}
\end{equation}
The reweighting reduces the load on the analytic visual axis
from 0.40 to 0.30 and redistributes 0.20 to the VLM axis.
 
\subsection{Iterative Refinement Protocol}
\label{sn:benchmark:loop}
 
Both pipelines follow the same iterative protocol, instantiated as
\texttt{IterativeCADEvaluator} in the accompanying notebooks.
Each problem is processed for up to $R=3$ refinement rounds indexed
$r \in \{0,1,2,3\}$.
Round~0 uses the initial system prompt (\texttt{SYSTEM\_INITIAL}) to
generate a first-pass JSON spec from the plain-text description.
Rounds $r\geq 1$ use a refinement system prompt
(\texttt{SYSTEM\_REFINE}) that injects three structured feedback
blocks into the user message:
\begin{enumerate}
  \item \textbf{Structural feedback}: schema errors, missing or
    incorrect features identified by the analytic scorer, and edge
    clearance violations.
  \item \textbf{Visual feedback}: the five sub-signals from the
    ray-trace critic (silhouette IoU, hole visibility, edge clearance
    visual score, aspect ratio, composite).
  \item \textbf{VLM semantic feedback} (\pipelineB{} only, rounds
    $r \leq 2$): the structured JSON response from Qwen2.5-VL-72B,
    including \texttt{missing\_features},
    \texttt{incorrect\_features}, \texttt{geometry\_issues}, and
    actionable \texttt{suggestions}.
\end{enumerate}
The LLM backend is queried via the Nebius AI Studio API
(OpenAI-compatible endpoint) at temperature 0.15 and a maximum of
2\,048 output tokens, with up to three retries per call with
exponential back-off.
The best-scoring spec across all rounds is retained as the final
output (\emph{best-of-$N$} selection); an early-exit criterion halts
refinement if the composite score reaches 0.92 or above.
The refinement prompt explicitly instructs the model to preserve all
holes, enforce a minimum edge clearance of $1.5\times$ the maximum
hole radius, correct schema errors, and add a fillet radius of 5--10\%
of the shorter principal dimension if absent.
 
\subsection{Models and Experimental Conditions}
\label{sn:benchmark:models}
 
Seven foundation models available on Nebius AI Studio are evaluated
under both pipelines on the same 97-problem benchmark:
Llama-3.3-70B-Instruct, DeepSeek-V3.2, INTELLECT-3,
Qwen3-235B-A22B-Instruct, Gemma-3-27B-IT, GLM-4.5, and MiniMax-M2.1.
All models are queried with the same temperature ($T=0.15$), token
budget (2\,048), and refinement protocol.
A cooldown of 5\,s is inserted between consecutive model runs to
avoid rate-limit interference.
Per-problem STL files of the best-round mesh are exported alongside
per-model JSON result files recording round-level trajectories for all
scoring axes.

\begin{figure}[ht]
  \centering
  \includegraphics[width=0.7\linewidth]{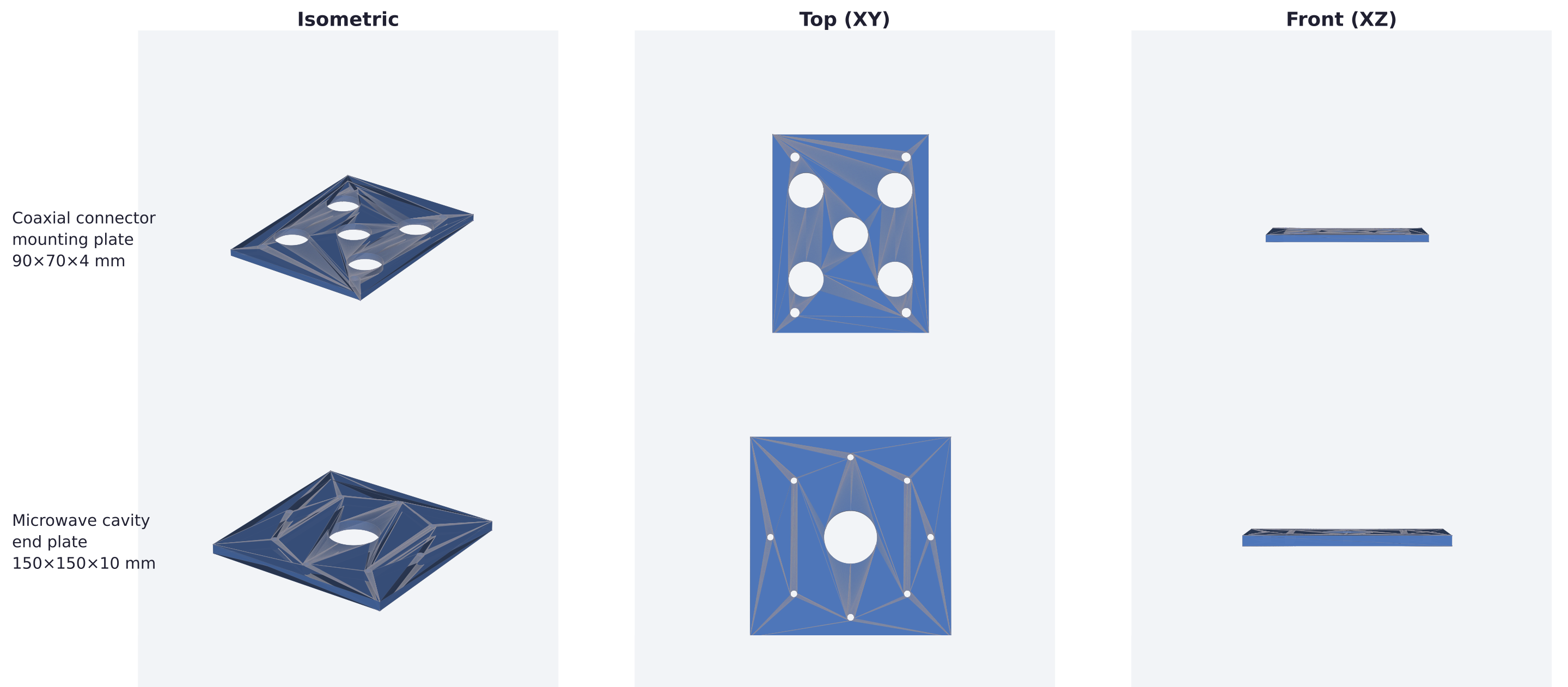}
  \caption{Example output meshes generated by DeepSeek-V3.2 under
           \pipelineB{} (best-round STL exports), rendered from three
           orthographic viewpoints: isometric, top (XY), and front (XZ).
           From top to bottom: a coaxial-connector mounting plate
           (90$\times$70$\times$4\,mm, 5 clearance holes) and a microwave-cavity
           end plate (150$\times$150$\times$10\,mm, central aperture with
           bolt circle). All meshes are watertight closed volumes; hole edges and
           Phong shading confirm correct feature geometry.}
  \label{fgr:mesh_gallery}
\end{figure}

\section{Critique Regimes: \pipelineA{} and \pipelineB{}}
\label{sn:pipelines}
 
 
Under \pipelineA{}, the visual feedback at each refinement round is
produced entirely by deterministic, differentiable geometric analysis
of the rendered bitmap,  no secondary neural model is involved.
After the geometry engine constructs the candidate mesh, two
orthographic views (top and isometric) are rendered at $256\times256$
pixels using a Matplotlib 3-D backend with Phong-like face colouring.
The top-view silhouette is extracted by thresholding non-white pixels
and used to compute: (i)~the \emph{silhouette~IoU}, defined as the
fraction of part pixels within the tight axis-aligned bounding box of
the projection, which serves as a proxy for mesh coherence and
dimensional completeness; (ii)~\emph{hole visibility}, the fraction of
declared holes whose circular projections remain detectable in the
top-view bitmap; (iii)~\emph{edge clearance}, the minimum normalised
distance from any visible hole projection to the plate boundary; and
(iv)~\emph{aspect-ratio conformance}, the relative deviation between
the rendered width-to-height ratio and the declared parameter ratio.
A fifth sub-signal, \emph{section score}, aggregates cross-sectional
consistency checks.
All five sub-signals are linearly combined into a single visual
composite (weights listed in \Cref{sn:benchmark:scoring}), which
feeds directly into the composite score $s_\text{overall}^\text{A}$
(\Cref{eq:score_v35}) and is reported verbatim to the model as
numerical feedback in the next-round refinement prompt.
The principal advantage of this regime is speed and determinism:
critique latency is sub-second, introduces no stochastic variance,
and does not require additional API calls.
 
 
\pipelineB{} augments the analytic critic with a Qwen2.5-VL-72B
semantic inspector deployed on Nebius AI Studio.
The renderer is upgraded to a full Phong-shading pipeline: per-face
normals are computed analytically, a Blinn-Phong model with ambient
$k_a=0.25$, diffuse $k_d=0.65$, and specular $k_s=0.10$ (shininess
8) is applied, and the light direction is co-aligned with the camera.
Three views are rendered at $384\times384$ pixels (top, isometric,
front) and base64-encoded before being sent to the VLM.
The VLM receives a structured multimodal prompt containing: the
original natural-language specification, the three rendered views, the
current JSON spec summary, and the automated visual metrics from the
analytic sub-pipeline.
It is instructed to respond with a single JSON object carrying
\texttt{semantic\_match} ($\in[0,1]$), \texttt{missing\_features},
\texttt{incorrect\_features}, \texttt{geometry\_issues},
\texttt{suggestions}, and \texttt{confidence}.
The \texttt{semantic\_match} value becomes $s_\text{vlm}$ and enters
the composite score (\Cref{eq:score_v40}); the textual fields are
concatenated into a third feedback block in the refinement prompt,
explicitly ranked above structural and analytic visual feedback in the
system-prompt priority order.
VLM critique is active only on refinement rounds $r \leq 2$ to limit
token expenditure; the total weight of the analytic visual axis is
reduced from 0.40 to 0.30 to make room for the 0.20 VLM axis.
A separate \texttt{vlm\_client} instance is initialised at VLM
temperature 0.05 to maximise determinism in the scoring output.

\section{Experimental Results}
\label{sn:results}
 
 
\Cref{t:summary} reports per-model mean scores across all 97
problems for both pipelines.
Under \pipelineA{}, four models,  DeepSeek-V3.2, Qwen3-235B-A22B,
Gemma-3-27B, and Llama-3.3-70B,  form a tight cluster in
$[0.885,\,0.890]$ with standard deviation $\sigma \leq 0.074$ and
uniform mesh success rates of 98.97\%.
The average number of refinement rounds consumed by this cluster
ranges from 3.58 to 3.65, indicating that the protocol rarely
triggers an early exit, with the best score typically arising from a
later round rather than round~0.
GLM-4.5 occupies a distinct second tier at 0.678 ($\sigma=0.305$),
while MiniMax-M2.1 and INTELLECT trail at 0.575 and 0.411
respectively, with high variance and sub-50\% mesh success rates.
 
Under \pipelineB{}, the top-four cluster shifts downward by
approximately 0.040 points uniformly, with overall scores in
$[0.842,\,0.850]$, reflecting the more demanding semantic assessment
pressure introduced by the VLM axis.
Gemma-3-27B is the sole model to achieve a mesh success rate of
\textbf{100\%} (97/97), and all 512 STL files produced by successful
evaluations are fully watertight and topologically valid closed
volumes.

Representative output geometries generated by DeepSeek-V3.2 are shown
in \Cref{fgr:mesh_gallery}, illustrating the range of hole patterns,
slot features, and dimensional proportions faithfully captured by the
pipeline across four distinct plate specifications.

The VLM axis rankings differ noticeably from the analytic-axis
rankings: DeepSeek-V3.2 leads on $s_\text{vlm}$ ($\mu=0.625$),
while Gemma-3-27B scores lowest among the top four ($\mu=0.511$),
suggesting that higher mesh reliability does not automatically
translate to stronger semantic correspondence as judged by the VLM
inspector.
 
\begin{table}[t]
\centering
\caption{Model performance summary. \pipelineA{} uses four scoring axes;
\pipelineB{} adds the VLM semantic axis. $\mu$ = mean, $\sigma$ = std,
SR = mesh success rate, $\bar{r}$ = average rounds.}
\label{t:summary}
\setlength{\tabcolsep}{4pt}
\footnotesize
\begin{tabular}{lcccc|ccccc}
\toprule
& \multicolumn{4}{c|}{\textbf{\pipelineA{}}}
& \multicolumn{5}{c}{\textbf{\pipelineB{}}} \\
\cmidrule(lr){2-5}\cmidrule(lr){6-10}
\textbf{Model} & $\mu$ & $\sigma$ & SR & $\bar{r}$
               & $\mu$ & $\sigma$ & SR & $\bar{r}$ & $s_\text{vlm}$ \\
\midrule
DeepSeek-V3.2         & 0.890 & 0.069 & 0.990 & 3.58
                      & 0.850 & 0.086 & 0.979 & 3.94 & 0.625 \\
Qwen3-235B-A22B       & 0.889 & 0.074 & 0.990 & 3.59
                      & 0.850 & 0.078 & 0.990 & 3.93 & 0.522 \\
Llama-3.3-70B         & 0.885 & 0.072 & 0.990 & 3.62
                      & 0.846 & 0.085 & 0.990 & 3.96 & 0.517 \\
Gemma-3-27B           & 0.885 & 0.068 & 0.990 & 3.65
                      & 0.842 & 0.069 & \textbf{1.000} & 3.95 & 0.511 \\
\midrule
GLM-4.5               & 0.678 & 0.305 & 0.691 & 3.68
                      & 0.561 & 0.276 & 0.546 & 3.95 & 0.564 \\
MiniMax-M2.1          & 0.575 & 0.286 & 0.495 & 3.90
                      & 0.552 & 0.245 & 0.485 & 3.99 & 0.575 \\
INTELLECT             & 0.411 & 0.290 & 0.268 & 3.96
                      & 0.401 & 0.271 & 0.289 & 4.00 & 0.607 \\
\bottomrule
\end{tabular}
\end{table}
 
\begin{figure}[ht]
  \centering
  \begin{subfigure}[t]{\linewidth}
    \includegraphics[width=\linewidth]{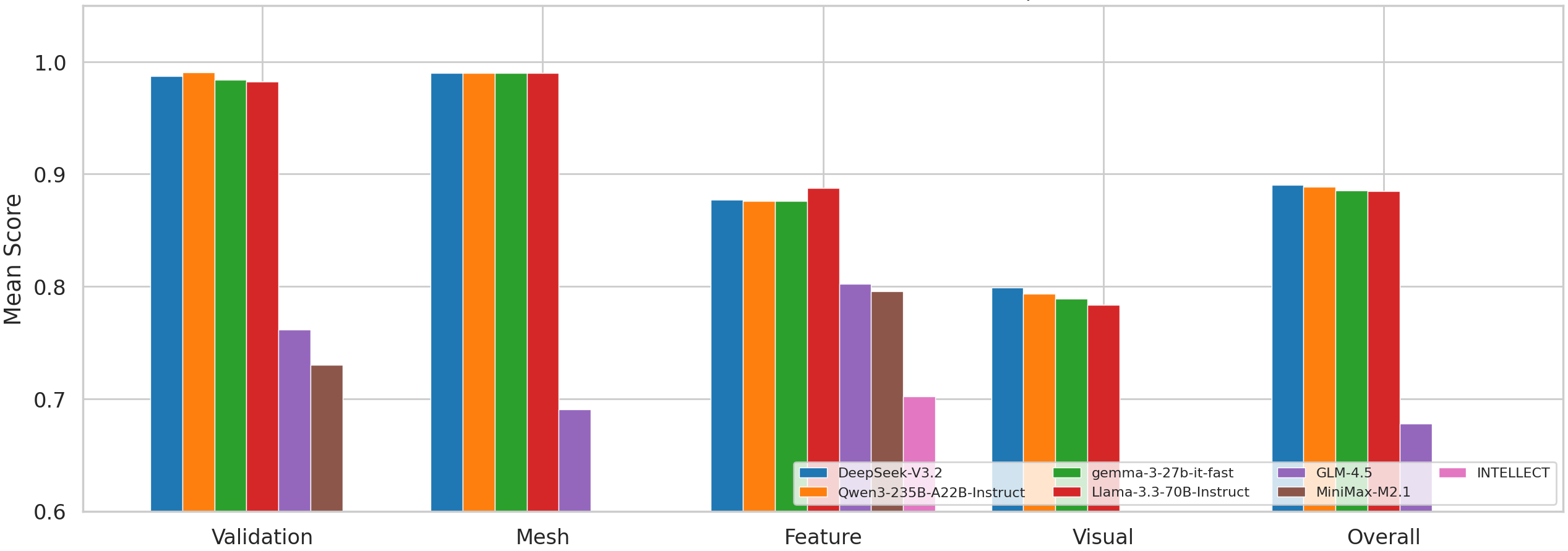}
    \caption{\pipelineA{} (4 axes).}
    \label{fgr:final_comparison_A}
  \end{subfigure}\\[4pt]
  \begin{subfigure}[t]{\linewidth}
    \includegraphics[width=\linewidth]{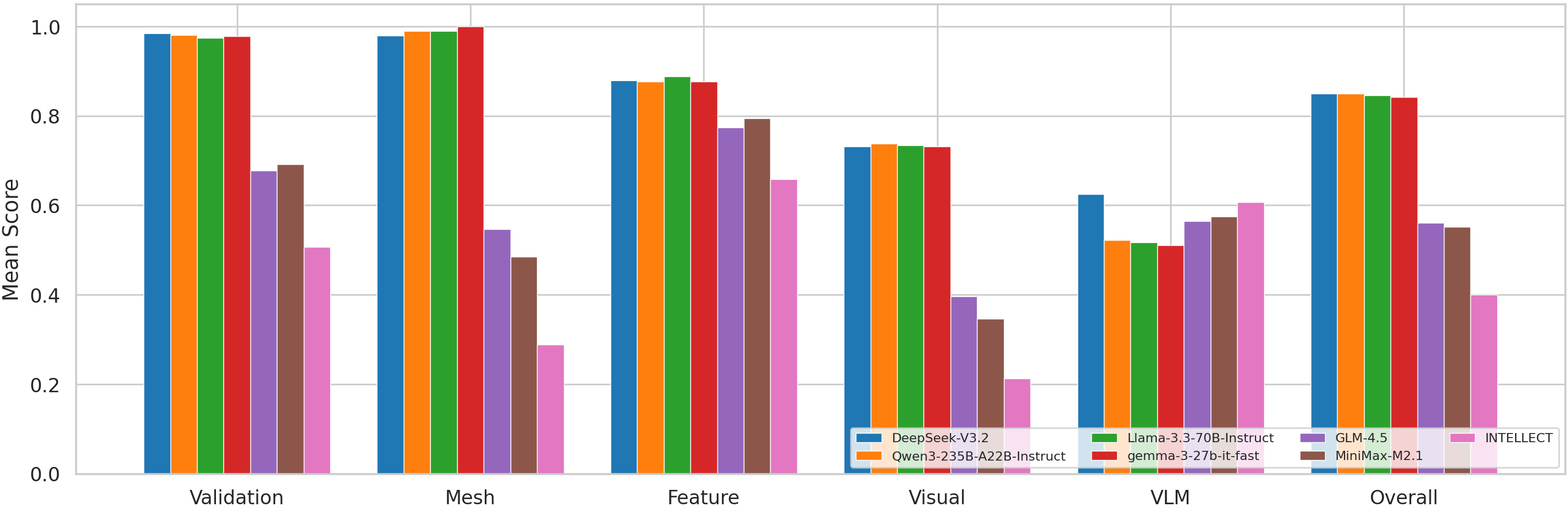}
    \caption{\pipelineB{} (5 axes, VLM added).}
    \label{fgr:final_comparison_B}
  \end{subfigure}
  \caption{Mean score per axis per model under both critique regimes
           (97 problems each). Comparing (a) and (b) shows that the VLM
           axis introduces new headroom and reorders models on the semantic
           dimension.}
  \label{fgr:final_comparison}
\end{figure}

\begin{figure}[H]
  \centering
  \begin{subfigure}[t]{0.48\linewidth}
    \includegraphics[width=\linewidth]{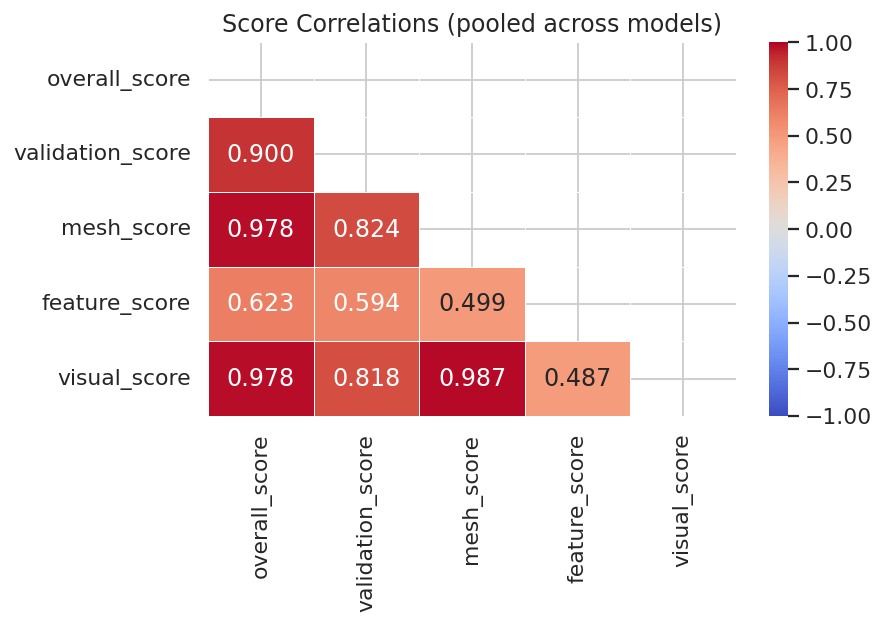}
    \caption{\pipelineA{} (4 axes).}
    \label{fgr:corr_A}
  \end{subfigure}\hfill
  \begin{subfigure}[t]{0.48\linewidth}
    \includegraphics[width=\linewidth]{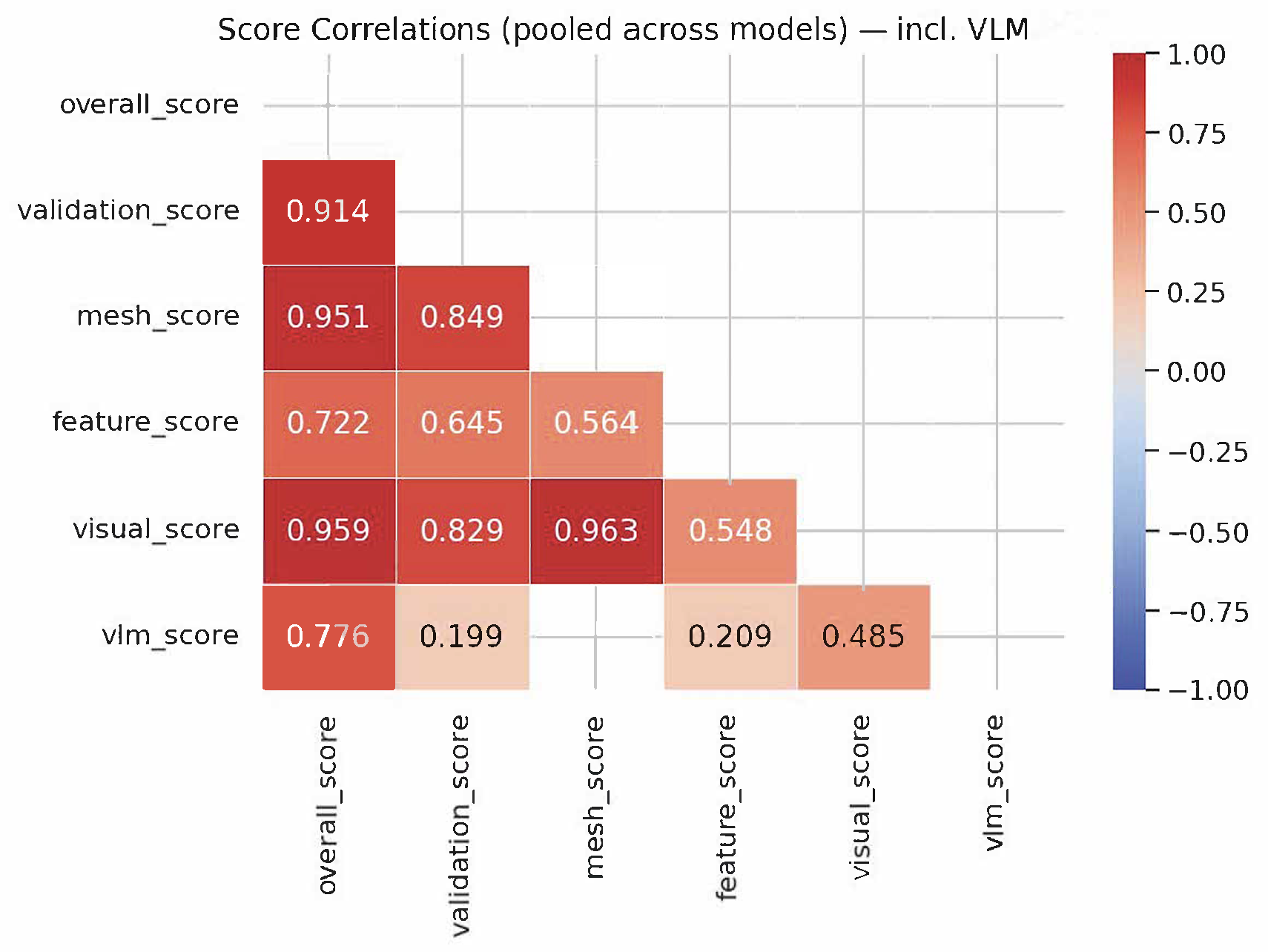}
    \caption{\pipelineB{} (5 axes).}
    \label{fgr:corr_B}
  \end{subfigure}
  \caption{Pearson correlation matrices (pooled across all models and
           problems) for the two critique regimes. Under \pipelineA{},
           mesh and visual scores are near-redundant; the VLM axis
           introduced in \pipelineB{} breaks this co-linearity and
           captures residual semantic variance invisible to analytic
           metrics.}
  \label{fgr:corr}
\end{figure}

 
\Cref{fgr:final_comparison} plots all five scoring axes per model
under \pipelineB{}.
The validation and mesh axes are near-ceiling ($>0.97$) for the
top-four cluster, confirming that schema conformance and geometric
validity are essentially saturated at this difficulty level.
The feature axis is more discriminating ($0.876$--$0.888$ for top
four; $0.658$--$0.794$ for the remaining three), and the visual axis
produces the largest absolute gap between tiers.
Pooled across models, the Pearson correlations with $s_\text{overall}$
are: $s_\text{vis}=0.959$, $s_\text{mesh}=0.951$,
$s_\text{val}=0.914$, $s_\text{vlm}=0.776$, $s_\text{feat}=0.722$.
The notably lower correlation of $s_\text{vlm}$ ($r=0.776$) compared
to $s_\text{vis}$ ($r=0.959$) confirms that the VLM critic captures a
partially orthogonal signal: it penalises semantic mismatches that
pass analytic visual filters, at the cost of introducing stochastic
scoring variance (\Cref{fgr:corr}).
 
Pairwise Bonferroni-corrected Mann-Whitney tests on
$s_\text{overall}$ show no statistically significant differences
within the top-four cluster (all adjusted $p=1.0$), while all
comparisons between any top-four model and the lower three are
significant ($p_\text{adj} \approx 0.0$).
GLM-4.5 vs.\ MiniMax-M2.1 are not significantly different
($p_\text{adj}=1.0$), but both differ significantly from INTELLECT
($p_\text{adj}=0.0026$ and $0.0002$ respectively).
 
 
\begin{figure}[ht]
  \centering
  \begin{subfigure}[t]{\linewidth}
    \includegraphics[width=\linewidth]{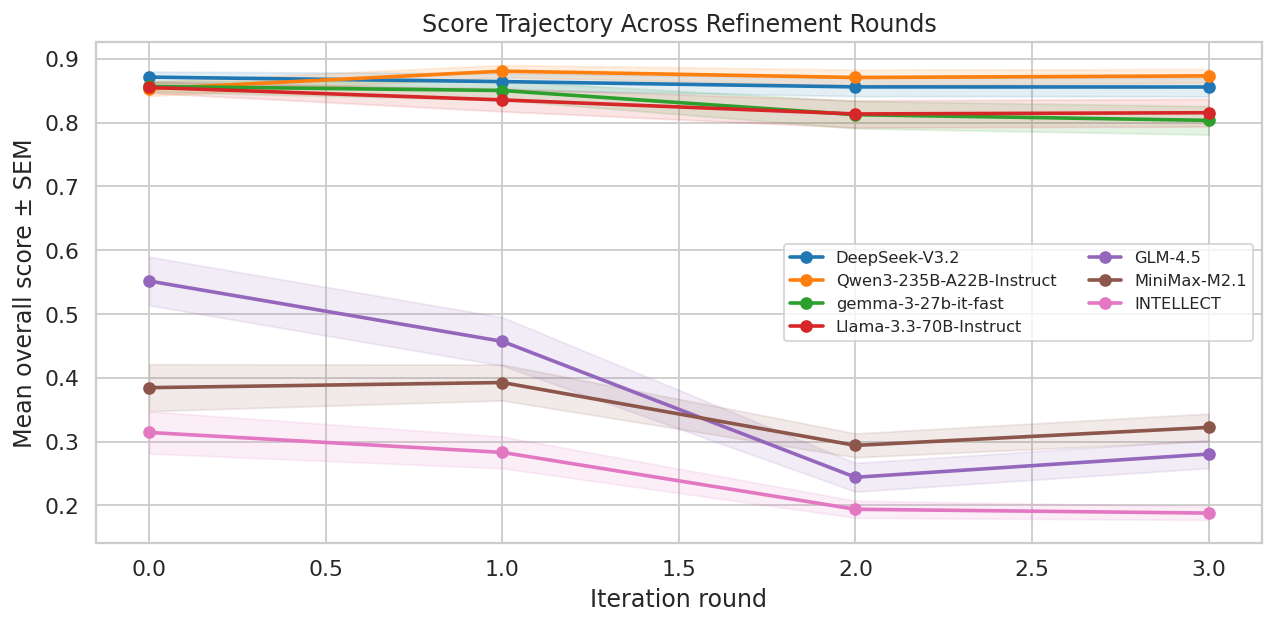}
    \caption{\pipelineA{}.}
    \label{fgr:trajectory_A}
  \end{subfigure}\\[4pt]
  \begin{subfigure}[t]{0.8\linewidth}
    \includegraphics[width=\linewidth]{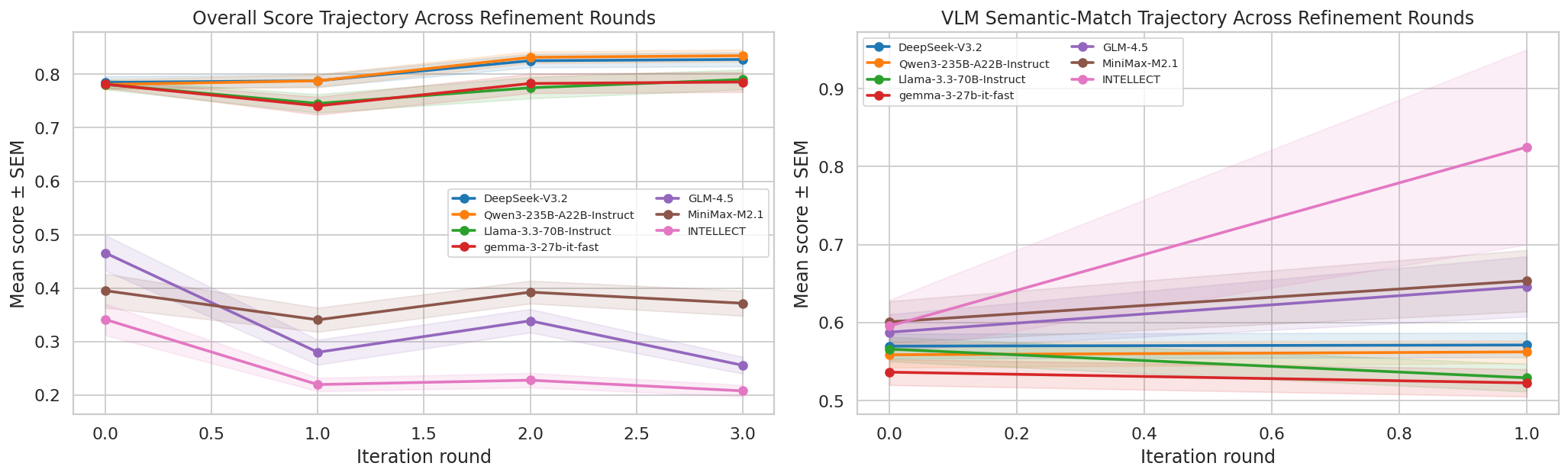}
    \caption{\pipelineB{}.}
    \label{fgr:trajectory_B}
  \end{subfigure}
  \caption{Per-round score trajectories ($\pm$SEM) under both critique
           regimes. The qualitative reversal between (a) and (b) for
           top-tier models demonstrates that VLM feedback sustains
           productive refinement beyond round~1 whereas analytic
           feedback saturates.}
  \label{fgr:trajectory}
\end{figure}

\Cref{fgr:trajectory} shows per-round mean overall score and VLM
semantic-match trajectories under \pipelineB{}.
For the top-four cluster, the overall score exhibits a shallow dip at
round~1 followed by monotonic recovery; 53--55\% of problems for
DeepSeek-V3.2 and Qwen3-235B-A22B achieve their best score at round~2,
with a further 31--38\% at round~3, confirming genuine multi-round
improvement under VLM-augmented critique.
 
The convergence regime under \pipelineA{} is qualitatively different.
Top-four models peak at round~1 for 62--83\% of problems (Qwen: 83\%,
Gemma: 71\%, Llama: 65\%, DeepSeek: 62\%) and their per-round
trajectories decline monotonically thereafter; median $\Delta$ for
the top cluster is 0.025 under \pipelineA{} versus $\sim$0.07 under
\pipelineB{}.
The analytic critic provides a strong one-shot corrective signal, 
schema errors and clearance violations repaired in round~1,  but its
numerical feedback is insufficiently expressive to sustain further
refinement, whereas the VLM's natural-language critique maintains
productive improvement across later rounds.
The correlation structure also shifts: under \pipelineA{}, both
$s_\text{mesh}$ and $s_\text{vis}$ reach $r=0.978$ with overall
(versus 0.951 and 0.959 under \pipelineB{}; see \Cref{fgr:corr}), indicating that without
a semantic axis the two analytic measures collapse to essentially
co-linear proxies.
 
For weaker models, the refinement benefit is larger in absolute terms
under both pipelines: median $\Delta$ reaches 0.19--0.21 for
GLM-4.5 and MiniMax-M2.1 under \pipelineB{}, and 0.16--0.18 under
\pipelineA{}, confirming that the feedback loop recovers a
substantially larger fraction of initially failed attempts for
lower-tier models.
The VLM trajectory (right panel, \Cref{fgr:trajectory}) reveals a
striking anomaly: INTELLECT's $s_\text{vlm}$ rises steeply from 0.60
to 0.82 across two VLM-active rounds, substantially outpacing all
other models on this axis despite ranking last on $s_\text{overall}$.
This divergence,  also visible in the raw VLM score distributions in
\Cref{fgr:final_comparison},  suggests that INTELLECT generates
geometries whose rendered appearance is rated favourably by the VLM
critic despite failing mesh validation, a form of \emph{visual--geometric
decoupling} that highlights the complementary nature of the two critique
modalities.

\section{Discussion}
\label{sn:discussion}
 
\paragraph{Critique modality determines optimal refinement depth.}
A cross-pipeline comparison reveals that the type of feedback signal,
not simply its presence, determines when iterative refinement stops
being beneficial.
Under \pipelineA{}, top-tier models saturate at round~1 and regress
on subsequent rounds; under \pipelineB{}, the same models continue
improving through rounds 2--3.
This suggests a general principle: analytic feedback corrects discrete
structural violations in a single pass, while semantic visual feedback
provides a richer, continuous quality gradient that supports deeper
search.
For production deployment, this implies that a two-stage strategy, 
analytic critique for rapid constraint satisfaction followed by
selective VLM invocation for high-value or ambiguous parts,  could
substantially reduce per-problem token expenditure without sacrificing
quality on the structurally simple majority.
 
\paragraph{Saturation of top-tier models under analytic critique.}
The near-identical performance of DeepSeek-V3.2, Qwen3-235B-A22B,
Llama-3.3-70B, and Gemma-3-27B under \pipelineA{}
($\Delta_\text{overall} < 0.005$ across the cluster, $p_\text{adj}=1.0$)
suggests that the four-axis analytic benchmark is essentially saturated
for state-of-the-art instruction-tuned models on this class of
canonical geometries.
The addition of the VLM semantic axis in \pipelineB{} restores
approximately 0.04 points of headroom and reorders the cluster
(Gemma-3-27B drops to rank~4 on VLM despite perfect mesh success),
demonstrating that semantic visual inspection surfaces genuine
residual variation that analytic metrics cannot capture.
 
\paragraph{Failure mode taxonomy.}
The lower tier exhibits two qualitatively distinct failure modes.
GLM-4.5 produces valid meshes in $\sim$55\% of cases but with high
variance ($\sigma=0.305$/$0.276$), suggesting intermittent schema
non-conformance or dimensional errors rather than systematic inability.
MiniMax-M2.1 and INTELLECT suffer more fundamental geometric failures:
sub-50\% mesh success rates indicate that the geometry engine
frequently raises exceptions on their output, pointing to malformed
JSON parameters (e.g.\ zero-size dimensions, inverted hole coordinates)
that schema validation cannot prevent.
INTELLECT's anomalously high VLM scores despite near-floor geometric
success expose a discrepancy between what the VLM perceives as
plausible geometry in a Phong-rendered image and what the analytic
pipeline enforces as valid topology.
 
\paragraph{Geometry-type difficulty and the cylinder gap.}
Across both pipelines, cylinder geometries score 0.04--0.07 points
below plates and L-brackets for the top-four models.
We attribute this to two compounding factors: (i)~the absence of
planar hole features removes the strongest visual discriminators
(hole visibility, edge clearance) from the cylinder score, lowering
the effective information content of the visual critique; and (ii)~the
VLM critic tends to rate cylindrical renders more conservatively,
possibly due to under-representation of rotating-symmetric industrial
components in its pretraining data.
This geometry-type gap suggests that future benchmark versions should
include flanged and threaded cylinder variants with richer feature sets
to provide cleaner diagnostic signals.
 
\paragraph{Industrial applicability.}
The results indicate that top-tier instruction-tuned foundation models
can generate dimensionally and topologically valid parametric parts
from natural-language descriptions with near-98\% mesh success rates
within three refinement iterations, at temperature 0.15 without
fine-tuning or retrieval augmentation.
The structured JSON representation and the multi-round critique
protocol are both amenable to integration into engineering PDM
workflows as a first-pass geometry generation module, with human
verification reserved for VLM-flagged semantic mismatches.
 
\section{Conclusion}
\label{sn:conclusion}
 
This chapter presented \framework{}, a multi-model text-to-CAD
evaluation framework studied under two critique regimes: \pipelineA{},
which employs analytic ray-trace visual metrics, and \pipelineB{},
which augments the analytic critic with a Qwen2.5-VL-72B semantic
inspector within the multi-round refinement loop.
Across a benchmark of 97 engineering design problems, four
state-of-the-art instruction-tuned models achieve near-identical
performance under analytic critique ($\mu_\text{overall}\approx0.887$,
mesh success 98.97\%), with Gemma-3-27B reaching 100\% watertight
mesh success under VLM-augmented critique.
The VLM axis surfaces a partially orthogonal quality signal
($r=0.776$ with overall score vs.\ $r=0.959$ for analytic visual),
restores meaningful score headroom saturated under analytic-only
evaluation, and exposes a visual--geometric decoupling phenomenon
in lower-tier models.
Iterative refinement provides consistent incremental benefit:
53--55\% of top-model problems attain their best score at round~2 or
later, and weaker models recover median $\Delta\approx0.19$ through
the feedback loop.
 
Future work will extend the benchmark to freeform surfaces, multi-body
assemblies, and manufacturing-specific constraints (GD\&T tolerances,
material specifications), and explore fine-tuning foundation models
directly on the structured feedback signal to narrow the gap between
top-tier and lower-tier models.
Integration with physics simulation backends,  replacing rendered
image critique with FEA-based structural validation,  represents a
natural pathway toward fully automated, simulation-in-the-loop
mechanical design generation.

\section*{Acknowledgments}
This research was supported by the LUXEMBOURG Institute of Science and Technology through the projects `ADIALab-MAST' and `LLMs4EU' (Grant Agreement No 101198470) and the BARCELONA Supercomputing Center through the project `TIFON' (File number MIG-20232039). Victoria Guill\'en would also like to thank Universidad Pontificia Comillas for the opportunity to participate in the international exchange program with Chung-Ang University, Seoul, Republic of Korea.

\section*{Code Availability}
The implementation of the \framework{} framework is publicly available at:
\url{https://github.com/drdecurto/LLMforge}.
The repository includes benchmark datasets, the geometry engine, both critique-regime notebooks (\pipelineA{} and \pipelineB{}), evaluation pipelines, per-model result files, and reproduction instructions.

\bibliographystyle{splncs04}
\bibliography{sample}

\end{document}